%% file: main.tex
\documentclass[11pt,a4paper]{article}

\usepackage{memoraxpaper}
\microtypesetup{expansion=false}
\usepackage{multirow}
\usepackage[
  backend=biber,
  style=authoryear-comp,
  natbib=true,
  maxcitenames=2,
  maxbibnames=12,
  uniquename=init,
  giveninits=true,
  doi=true,
  url=true,
  isbn=false,
  eprint=true
]{biblatex}
\AtBeginBibliography{\raggedright\sloppy}

\newcommand{\method}{\textsc{GraphMemix}}

\renewenvironment{abstract}{\begin{reportabstract}}{\end{reportabstract}}
\tcbset{
  graphprompt/.style={
    colback=MemoraXPaper,
    colframe=MemoraXRule,
    arc=2pt,
    fonttitle=\bfseries\ttfamily
  }
}
\reporttitle{GraphMemix: Query-Aware Evidence Forests for Long-Term Multimodal Agent Memory}
\reportsubtitle{}
\reportcategory{}
\reportrunningtitle{GraphMemix}
\reportauthors{Geng Li\texorpdfstring{\textsuperscript{1}}{}, Yuhao Wang\texorpdfstring{\textsuperscript{1}}{}, Dong Li\texorpdfstring{\textsuperscript{2}}{}, Jianye Hao\texorpdfstring{\textsuperscript{2}}{}, Yuxin Peng\texorpdfstring{\textsuperscript{1}}{}}
\reportaffiliations{\textsuperscript{1}Wangxuan Institute of Computer Technology, Peking University\\\textsuperscript{2}MemoraX AI}
\reportversion{}
\reportrunningversion{Technical Report}
\reportdate{}
\reportcontact{Correspondence to: Yuxin Peng \href{mailto:pengyuxin@pku.edu.cn}{\texttt{<pengyuxin@pku.edu.cn>}}}
\reportfooter{MemoraX AI Research · GraphMemix}
\reportlogo{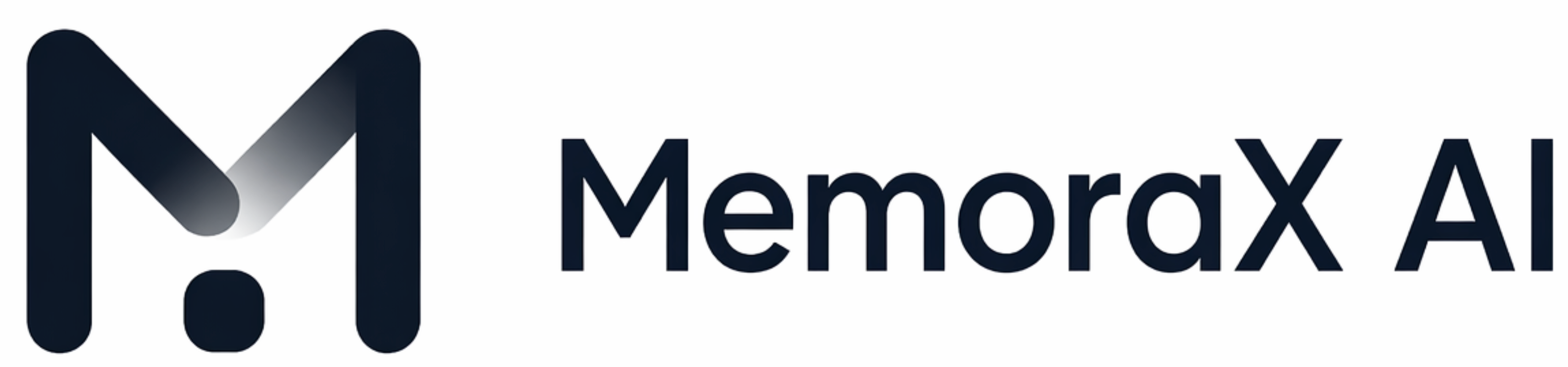}
\reportsubject{Query-aware evidence forests for long-term multimodal agent memory}
\reportkeywords{multimodal agent memory, evidence retrieval, graph optimization, long-term memory}

\begin{document}

\makememoraxpapertitle
\input{sections/abstract}
\printreportkeywords

\input{sections/introduction}
\input{sections/related_work}
\input{sections/method}
\input{sections/experiments}
\input{sections/conclusion}

\section{References}
\printbibliography[heading=none]

\appendix
\input{sections/appendix}

\end{document}

%% file: sections/abstract.tex
\begin{abstract}
Organizing long-term memory for multimodal agents remains challenging because existing methods either suffer from expensive question-agnostic offline summaries or naive embedding similarity matching that introduces incomplete and redundant context. To address these issues, we propose \method{}, a \textbf{combinatorial-optimization graph memory framework} that \textbf{models} memory organization as query-aware evidence-forest construction. Specifically, our method consists of three key components:
(1) \textbf{candidate graph construction}, which expands multi-view seed memories through schema and semantic relations to acquire query-aware original context;
(2) \textbf{evidence utility and activation costs}, which decouples direct memory support from anchor-conditioned relation verification to suppress redundant or conflicting information; and
(3) \textbf{forest optimization}, which jointly selects a forest-format memory context under a maximum evidence budget and its reliable relational structure.
By organizing memory into a query-relevant subgraph, the method avoids substantial lifecycle cost and recovers low-similarity complementary evidence. Experimental results across four long-term multimodal memory benchmarks demonstrate significant improvements with different foundation models and establish a new Pareto frontier between accuracy and lifecycle cost.
\end{abstract}

%% file: sections/introduction.tex
\section{Introduction}

Memory serves as a fundamental component for Large Vision-Language Model (LVLM)-based agents \cite{feng2026m2a,wang2025mirix} 
to maintain consistency across long-horizon multimodal interactions and incorporate new information \cite{ren2026memlens,bei2026memgallery}. 
Especially, for personal assistant agents \cite{wang2025mirix,feng2026m2a}, the memory system must encompass not only 
historical user-agent interactions \cite{bei2026memgallery,feng2026m2a}, 
but also vast volumes of continuously updated, 
user-centric multimodal data, such as photo albums, emails, personal chat logs, 
work documents and etc. \cite{mei2026atm,wang2025mirix}. 
This scale and heterogeneity make it difficult for agent memory to retrieve
precise, query-relevant context
\cite{jiang2025memoryqa,du2026memguide,li2026memreranker}.

Existing methods mainly follow two routes. The first extends text-based agent
memory by converting history into reusable summaries, typed fields,
or linked notes before the future question is known. MIRIX
\cite{wang2025mirix} partitions experience into episodic, semantic,
procedural and knowledge-vault memories. SGM
\cite{mei2026atm} maps source-specific image, video, and email content into a
shared record with meta fields. A-MEM
\cite{xu2025amem} constructs structured notes and updates their links;
AUGUSTUS \cite{jain2025augustus} and M$^2$A \cite{feng2026m2a} further connect
semantic memories to original records. These \emph{question-agnostic} memory methods compress history to build a general-purpose
representation before queries arrive.
However, because no future question is considered, the whole user history are processed by default to avoid omission, creating a large cold-start and
continuing update cost. More
fundamentally, question-agnostic compression may omit visual attribute, state transition, or local
context that later becomes decisive \cite{guo2026memeye,ren2026memlens}. 
The second route is \emph{multimodal RAG}. MuRAG \cite{chen2022murag} retrieves
directly from a corpus of images and text, while Pensieve
\cite{jiang2025memoryqa} combines images, captions, OCR, spatio-temporal cues,
and multimodal similarity to search personal visual experience. 
These methods preserve native visual access and cost low in processing. However, ranking records primarily by 
similarity matching leads them to overselect near-duplicates while omitting dissimilar 
replies or state updates whose value emerges only in 
combination with other records. Prior work shows that answering a question can require multiple
complementary memories, while semantic similarity alone does not determine
their joint evidential utility \cite{jiang2025memoryqa,du2026memguide}.



To address these limitations, we propose \method, a combinatorial-optimization graph memory framework that reconstructs necessary
evidence forest from large-scale memory archives after the question arrives.
Fundamentally, \method{} formulates memory selection as query-conditioned
evidence-forest optimization: nodes capture the direct utility of individual
memories, while edges capture their incremental value when conditioned on an
anchor. To instantiate this objective, \method{} starts from multi-view matching
anchors and expands through schema and semantic edges to reconstruct a bounded,
query-relevant candidate subgraph. A node verifier estimates direct support,
while an Evidence-Chain Verifier estimates whether each anchor-conditioned
relation contributes complementary information rather than redundancy or
conflict. The resulting objective jointly selects relevant memories and
reliable relations, recovering low-similarity context while suppressing
uncertain expansion. For any fixed node set, its optimal maximum-weight forest
is obtained exactly by Kruskal's algorithm. 
Experiments on four long-term multimodal memory benchmarks and two reader
families support this design. With Qwen3-VL, \method{} improves Judge Accuracy
over the strongest public baseline on every benchmark and raises the
four-dataset macro-average by 11.75 percentage points, from 49.80\% to 61.55\%.
In terms of efficiency, it achieves a lifecycle-cost Pareto frontier compared with existing approaches, demonstrating that query-local semantic organization boosts answer quality without incurring history-wide generative preprocessing overhead.



%% file: sections/related_work.tex
\section{Related Work}

\subsection{Long-Term Multimodal Agent Memory}

Long-term memory enables agents to preserve and reuse interaction history
beyond a finite context. Generative Agents \cite{park2023generative} write
observations into a memory stream and synthesize reflections; MemGPT
\cite{packer2023memgpt} controls movement across hierarchical storage; A-MEM
\cite{xu2025amem} turns new experiences into structured notes and updates
links to existing memory. These
works establish persistent writing, organization, and controlled retrieval, but
primarily operate on textual experiences or textualized memory units.
Multimodal memory systems make text and images jointly searchable. MuRAG
\cite{chen2022murag} uses non-parametric multimodal memory for
knowledge-intensive QA; Pensieve \cite{jiang2025memoryqa} combines captions,
OCR, spatiotemporal metadata, and multimodal similarity for personal visual
experience; MIRIX \cite{wang2025mirix} and SGM \cite{mei2026atm} organize
history as typed memories or a unified schema. AUGUSTUS
\cite{jain2025augustus} and M$^2$A \cite{feng2026m2a} connect semantic indices
or editable memories to logs that retain the original media. 
Such approaches either demand lengthy preprocessing or lose structural context of stored memories.
\method{} focuses on reconstructing the evidence subgraph conditioned on each query, thereby reducing lifecycle computational overhead and improving the completeness of memory context.

\begin{figure}[t]
  \centering
  \includegraphics[width=\textwidth]{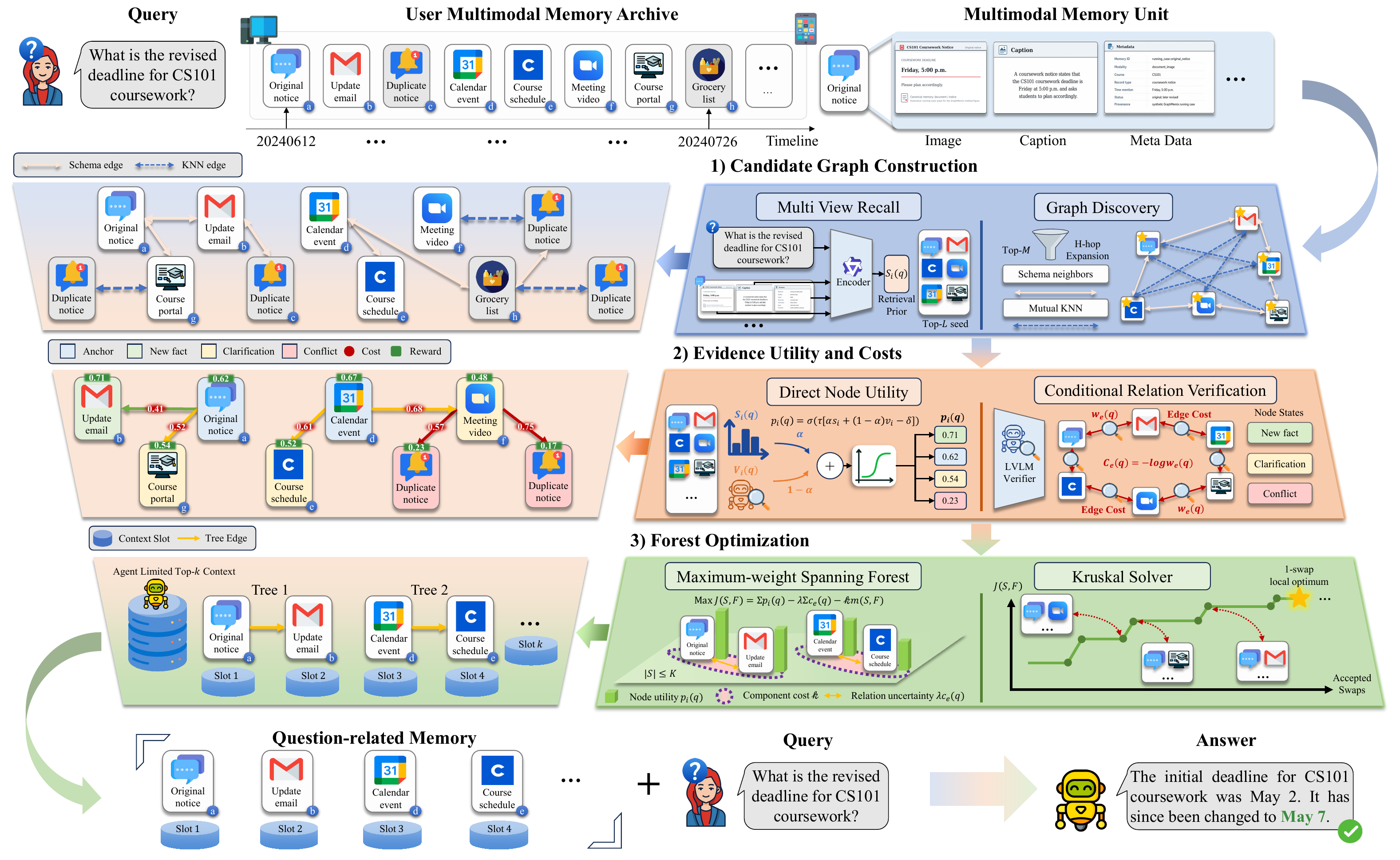}
  \caption{Overview of \method{} in three stages. (1) \emph{Candidate graph
  construction} uses multi-view retrieval to identify seeds and bounded
  relation expansion to expose their local context. (2) \emph{Evidence utility
  and costs} combines direct node support with ECV-validated,
  anchor-conditioned incremental relations. (3) \emph{Forest optimization}
  jointly selects nodes and trusted edges under a maximum reader budget, then
  deterministically serializes the resulting evidence forest.}
  \label{fig:method}
\end{figure}

\subsection{Query-Conditioned Evidence Retrieval}

Query-conditioned retrieval adapts memory access to the information need
expressed by the current question. MemGuide \cite{du2026memguide} retrieves
intent-aligned memories and then filters them according to missing information,
while MemReranker \cite{li2026memreranker} targets temporal constraints, causal
relations, and multi-turn coreference that challenge generic relevance models.
In general retrieval, RankGPT \cite{sun2023rankgpt} shows that listwise
comparison with a generative LLM can outperform independent similarity
scoring. These methods primarily produce
query--memory relevance scores or rankings; candidate--candidate relations are
not explicit decision variables.
Multimodal RAG further asks whether retrieved content genuinely supports
generation. RagVL \cite{chen2024ragvl} uses an LVLM to rerank retrieved images,
MEG-RAG \cite{wang2026megrag} trains a multimodal evidence reranker around the
semantic core of an answer. 
These works generally ignore inter-evidence correlations and are vulnerable to redundant or contradictory evidence.
\method{} expands query-conditioned scoring to complete context construction, jointly modeling evidence utility and valid inter-evidence links while discarding redundant similar memories.

%% file: sections/method.tex
\section{Method}
\label{sec:method}

We propose \method{}, which formulates long-term multimodal memory organization as
a query-conditioned forest optimization problem over a bounded candidate graph.


\subsection{Problem Formulation}

Let $\mathcal V=\{e_i\}_{i=1}^{N}$ be user multimodal memory archive containing $N$
historical records, and let $q=(q^{\mathrm{text}},q^{\mathrm{vis}})$ be a
question with visual input available. Each $e_i$ may contain text, images,
video, and derived representations. An evidence selector $\Pi$ chooses an
evidence set of at most $K$ records and organizes it into an ordered sequence
$O_q$. A frozen
multimodal generator $\mathcal M$ then produces the answer:
\begin{align}
(S_q,O_q)&=\Pi(q,\mathcal V;K),\qquad
S_q\subseteq\mathcal V,\quad |S_q|\leq K,\nonumber\\
\hat y&=\mathcal M(q,O_q).
\label{eq:task}
\end{align}
Our objective is to design $\Pi$ such that the selected evidence provide complete context to improve the
correctness of $\hat y$.

\subsection{Query-Conditioned Evidence Forest}

A direct implementation of $\Pi$ is ranking memories independently by
query-memory similarity. It can waste slots on duplicates and miss a response,
subsequent state, or referential context whose wording differs from the
question. Conversely, inserting every connected record introduces irrelevant
context. We instead represent candidate memories as nodes and possible
contextual dependencies as edges. For selected nodes $S$ and an acyclic edge
set $F$, we write the selection objective as
\begin{equation}
\max_{S,F}\;
\mathcal U_{\mathrm{node}}(S\mid q)
-\mathcal C_{\mathrm{edge}}(F\mid q)
-\mathcal C_{\mathrm{open}}(S,F),
\label{eq:abstract_obj}
\end{equation}
where the terms measure direct evidence utility $U_{\mathrm{node}}$, uncertainty in expanding a
relation $C_{\mathrm{edge}}$, and the cost of opening independent evidence chains $C_{\mathrm{open}}$. The remainder
of this section constructs and solves this objective.

\subsection{Candidate Graph Construction}

\paragraph{Multi-view retrieval.}
The same memory may be retrieved through different signals: an original image
for objects and scenes, a caption for events and actions, OCR for printed
text, or a video frame for visual state. Let $\mathcal U_i$ be the retrieval
views of $e_i$. Its initial score is
\begin{equation}
s_i(q)=\max_{u\in\mathcal U_i}
\operatorname{sim}\!\left(h(q),h(u)\right),
\label{eq:multiview}
\end{equation}
where $h$ is a shared multimodal encoder. Max pooling only provides multiple
routes into the candidate set. 

\paragraph{Query-relevant subgraph discovery.}
We use two complementary relation sources. Schema relations encode observable
interaction structure, such as records from the same session or location.
Semantic relations connect memories with mutually similar 
representations. Let $\mathcal T_{\mathrm{schema}}$ be the schema relation
types, $\phi_r(e_i,e_j)$ indicate whether relation $r$ holds, and
$\mathcal N_k(i)$ denote the $k$ nearest neighbors of $i$. We define
\begin{align}
E_{\mathrm{schema}}
&=\{(i,j):\exists r\in\mathcal T_{\mathrm{schema}},
\ \phi_r(e_i,e_j)=1\},\nonumber\\
E_{\mathrm{sem}}
&=\{(i,j):i\in\mathcal N_k(j),\ j\in\mathcal N_k(i)\},
\label{eq:discovery_edges}
\end{align}
and the discovery graph
$G^{\mathrm{disc}}=(\mathcal V,E_{\mathrm{schema}}\cup E_{\mathrm{sem}})$.
Starting from the top-$L$ multi-view memories
$R_q^{(L)}=\operatorname{Top}_L(\mathcal V;s_i(q))$, we collect nodes reachable
within $H$ hops and retain at most $M$ candidates:
\begin{equation}
\widetilde C_q=R_q^{(L)}\cup
N_{\le H}^{G^{\mathrm{disc}}}(R_q^{(L)}),\qquad
C_q=\operatorname{Top}_M(\widetilde C_q).
\label{eq:candidate}
\end{equation}
The ranking combines retrieval order and relation-expansion order. This step
preserves local context while bounding all subsequent semantic reasoning.

\subsection{Evidence Utility and Activation Costs}

\paragraph{Direct node utility.}
Retrieval similarity is a stable global prior, but is insufficient for
negation, temporal change, visual reference, or memories that share a topic but
imply different answers. A listwise node verifier reads $q$ and $C_q$ in one
call and outputs $v_i(q)$, the degree to which $e_i$ independently supports the
question. Listwise input places candidates on a common semantic scale. We fuse
the normalized verifier score $\widetilde v_i$ with retrieval:
\begin{align}
p_i(q)&=\sigma\!\left(\tau[
\alpha s_i(q)+(1-\alpha)\widetilde v_i(q)-\delta]\right),\nonumber\\
\mathcal U_{\mathrm{node}}(S\mid q)&=\sum_{i\in S}p_i(q),
\label{eq:node_utility}
\end{align}
where $\alpha$ balances the retrieval prior and conditional judgment, while
$\tau$ and $\delta$ control scale and offset.

\paragraph{Anchor-conditioned relation verification.}
Node verification asks whether a memory is useful alone, but cannot distinguish
redundancy from complementarity with an already retrieved memory. We take a
high-confidence anchor set $A_q$ from the initial retrieval and inspect only
schema edges between anchors and candidates:
\begin{equation}
E_q^{\mathrm{elig}}=\{(a,i)\in E_{\mathrm{schema}}:
a\in A_q,\ i\in C_q\}.
\label{eq:eligible}
\end{equation}
The Evidence-Chain Verifier (ECV) reads these pairs in one listwise call. For
each candidate, it chooses the anchor that best explains the candidate's
incremental role, predicts an incremental-support score
$s_{ai}^{\mathrm{inc}}(q)$, and assigns one of six roles:
\texttt{new\_fact}, \texttt{clarification}, \texttt{corroboration},
\texttt{redundant}, \texttt{conflict}, or \texttt{irrelevant}. Only the best
anchor edge with a positive score and one of the first three roles is retained,
forming
\begin{equation}
G_q^{\mathrm{trust}}=(C_q,E_q^{\mathrm{trust}}),\qquad
E_q^{\mathrm{trust}}\subseteq E_q^{\mathrm{elig}}.
\label{eq:trust}
\end{equation}
For a retained edge $(a,i)$, its reliability and uncertainty cost are
\begin{equation}
w_{ai}(q)=w_{ai}^{\mathrm{schema}}
\frac{s_{ai}^{\mathrm{inc}}(q)}{s_{\max}},\qquad
c_{ai}(q)=-\log w_{ai}(q),
\label{eq:edge_cost}
\end{equation}
where $w_{ai}^{\mathrm{schema}}$ is a schema-specific reliability ceiling.
Accordingly,
$\mathcal C_{\mathrm{edge}}(F\mid q)=
\lambda\sum_{e\in F}c_e(q)$.
Role gating determines whether an edge carries positive incremental semantics,
the continuous score then controls its credit. A related but redundant or
conflicting memory therefore receives no structural reward.

\paragraph{Independent-chain cost.}
Let $m(S,F)$ be the number of connected components in the forest $(S,F)$.
Every component is an independently interpreted evidence chain, so we define
\begin{equation}
\mathcal C_{\mathrm{open}}(S,F)=\kappa\,m(S,F),
\label{eq:open_cost}
\end{equation}
where $\kappa\geq0$ balances independent evidence against contextual
completion of an existing event.

\input{tables/main_results_qwen}
\input{tables/main_results_gemma}

\begin{figure}[t]
  \centering
  \includegraphics[width=\columnwidth]{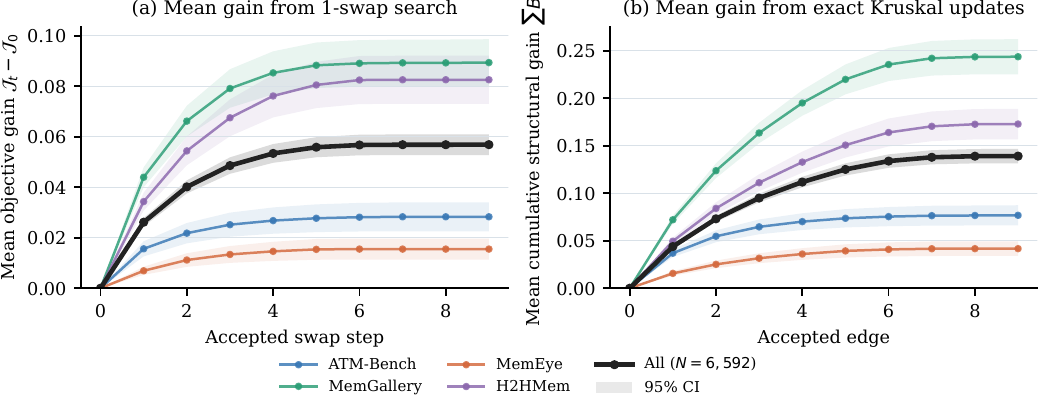}
  \caption{Optimization behavior over all 6,592 evaluation questions.
  (a) Mean improvement in the complete forest objective after accepted
  1-swap updates. (b) Mean cumulative structural gain from accepted edges
  during the exact Kruskal update. Shaded regions are 95\% confidence intervals.}
  \label{fig:optimization_trace}
\end{figure}

\begin{figure}[t]
  \centering
  \includegraphics[width=\textwidth]{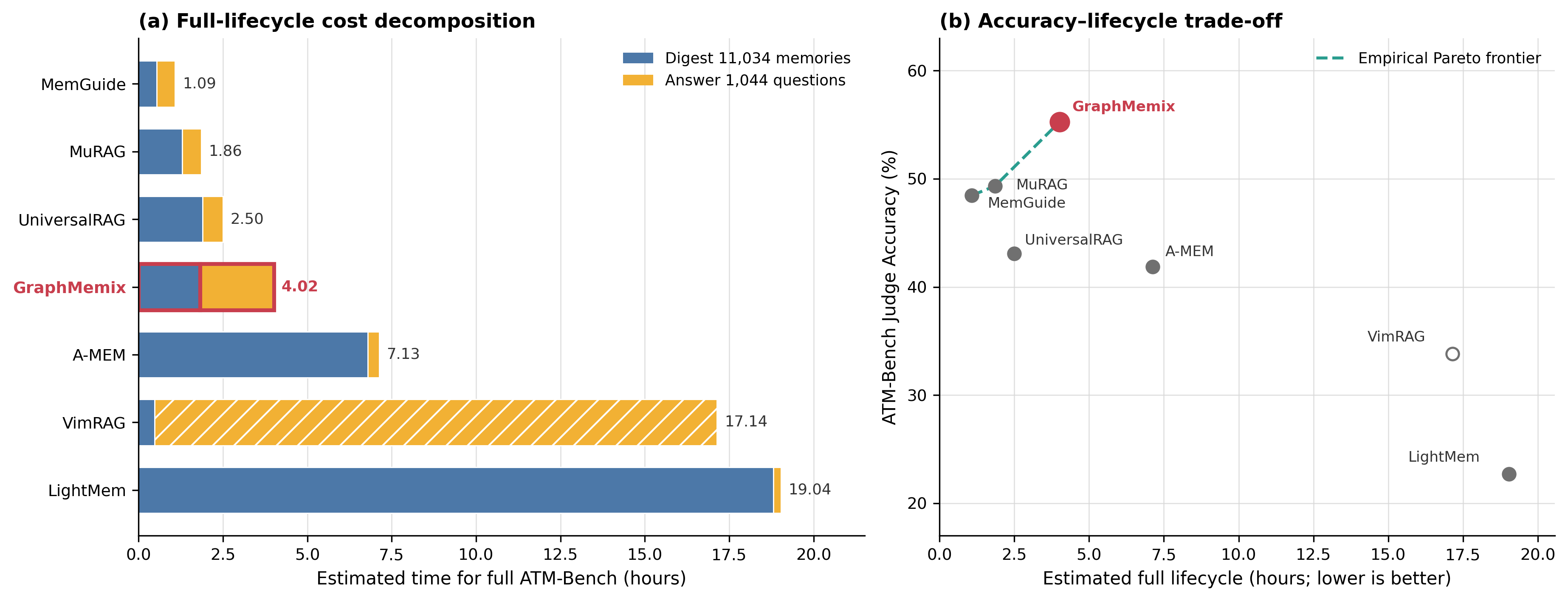}
  \caption{Full-lifecycle cost and accuracy on ATM-Bench. (a) Estimated time to
  digest all 11,034 memories once and answer all 1,044 questions once.
  (b) Judge Accuracy versus the same lifecycle time. \method{} lies on the
  empirical Pareto frontier.}
  \label{fig:lifecycle}
\end{figure}

\subsection{Forest Optimization}

Substituting Equations~\eqref{eq:node_utility}--\eqref{eq:open_cost} into
Equation~\eqref{eq:abstract_obj} gives
\begin{equation}
\begin{aligned}
(S_q^\star,F_q^\star)=\arg\max_{S,F}\;&
\sum_{i\in S}p_i(q)-\lambda\sum_{e\in F}c_e(q)
-\kappa m(S,F)\\
\text{s.t.}\quad&
S\subseteq C_q,\quad |S|\leq K,\\
&F\subseteq E_q^{\mathrm{trust}}[S].
\end{aligned}
\label{eq:forest}
\end{equation}
Since edge costs are nonnegative and removing an edge from any cycle cannot decrease the objective, an
optimal $F$ can therefore be chosen as a forest. Hence,
$m(S,F)=|S|-|F|$. Defining the
structural gain $B_e(q)=\kappa-\lambda c_e(q)$ gives the equivalent utility
\begin{equation}
\sum_{i\in S}\bigl(p_i(q)-\kappa\bigr)
+\sum_{e\in F}B_e(q).
\label{eq:reduced}
\end{equation}
This form explains adaptive cardinality without introducing a separate
per-node penalty. An isolated memory contributes $p_i-\kappa$ and is retained
only if it can justify opening an independent evidence chain. If a memory
joins an existing component through edge $e$, the connection recovers one
opening cost and its marginal contribution becomes $p_i-\lambda c_e$.
Trusted complementary evidence is therefore protected, whereas low-utility
isolated evidence can be removed.

\paragraph{Bounded exact cardinality refinement.}
Joint optimization over all $M$ candidates is combinatorial. We use a
deterministic two-stage solver that preserves the recall behavior of the
bounded candidate graph. First, the fixed-cardinality forest solver produces
a $K$-node proposal $\bar S_q\subseteq C_q$: it initializes from the top-$K$
node utilities, evaluates one selected--unselected swap at a time, and uses
Kruskal's algorithm to recompute the exact maximum-weight forest for every
proposed node set. This stage terminates at a deterministic 1-swap local
optimum.
Second, we solve the variable-cardinality problem exactly inside the frozen
proposal:
\begin{align}
F_S^\star
&=\arg\max_{\substack{F\subseteq E_q^{\mathrm{trust}}[S]}}
\sum_{e\in F}B_e(q),\nonumber\\
(S_q^\star,F_q^\star)
&=\arg\max_{\varnothing\neq S\subseteq\bar S_q}
\left[
\sum_{i\in S}\bigl(p_i(q)-\kappa\bigr)
+\sum_{e\in F_S^\star}B_e(q)
\right].
\label{eq:exact_refinement}
\end{align}
For each subset, Kruskal applied to positive-gain edges gives
$F_S^\star$ exactly. We break
objective ties by preferring fewer memories and then earlier proposal ranks.
The returned solution is consequently globally optimal over all nonempty
subsets of $\bar S_q$; Evidence ordering details are provided in the
supplementary material.

%% file: tables/main_results_qwen.tex
\begin{table}[t]
\centering
\normalsize
\setlength{\tabcolsep}{4pt}
\resizebox{\linewidth}{!}{%
\begin{tabular}{llrrrrrrrrr}
\toprule
& & \multicolumn{2}{c}{ATM} & \multicolumn{2}{c}{Mem-Gallery} &
\multicolumn{2}{c}{MemEye} & \multicolumn{2}{c}{H2HMem} & Avg. \\
\cmidrule(lr){3-4}\cmidrule(lr){5-6}\cmidrule(lr){7-8}\cmidrule(lr){9-10}
Method & Venue & J.Acc. & EM & J.Acc. & EM & J.Acc. & M.EM &
J.Acc. & Recall & J.Acc. \\
\midrule
A-MEM        & NeurIPS'25 & 41.86 & 40.80 & 52.89 & 32.61 & 37.09 & 40.57 & 39.94 & 40.55 & 42.94 \\
UniversalRAG & ACL'26     & 43.10 & 41.19 & \underline{63.76} & 32.50 & \underline{47.98} & \underline{49.33} & \underline{44.36} & \underline{45.04} & \underline{49.80} \\
MemGuide     & AAAI'26    & \underline{48.47} & \underline{47.80} & 48.10 & \underline{32.73} & 41.40 & 44.61 & 39.94 & 35.81 & 44.48 \\
LightMem     & ICLR'26    & 22.70 & 24.52 & 38.28 & 23.20 & 36.23 & 39.76 & 31.04 & 27.82 & 32.06 \\
VimRAG       & ICML'26    & 33.81 & 32.76 & 48.33 & 16.13 & 28.25 & 28.23 & 17.89 & 17.01 & 32.07 \\
\midrule
\method{}    & Ours & \textbf{55.27} & \textbf{53.83} &
\textbf{76.33} & \textbf{36.76} &
\textbf{53.64} & \textbf{54.25} &
\textbf{60.96} & \textbf{54.70} & \textbf{61.55} \\
\textcolor{green!50!black}{Gain over 2nd} & \textcolor{green!50!black}{--} &
\textcolor{green!50!black}{+6.80} & \textcolor{green!50!black}{+6.03} &
\textcolor{green!50!black}{+12.57} & \textcolor{green!50!black}{+4.03} &
\textcolor{green!50!black}{+5.66} & \textcolor{green!50!black}{+4.92} &
\textcolor{green!50!black}{+16.60} & \textcolor{green!50!black}{+9.66} &
\textcolor{green!50!black}{+11.75} \\
\bottomrule
\end{tabular}%
}
\caption{End-to-end results with Qwen3-VL-8B-Instruct, judged by GPT-5-mini. J.Acc.\ denotes Judge Accuracy (\%).
Each benchmark additionally reports its representative native metric. Avg.\ is
the macro-average over the four Judge Accuracies.}
\label{tab:main}
\end{table}

%% file: tables/main_results_gemma.tex
\begin{table}[t]
\centering
\normalsize
\setlength{\tabcolsep}{4pt}
\resizebox{\linewidth}{!}{%
\begin{tabular}{llrrrrrrrrr}
\toprule
& & \multicolumn{2}{c}{ATM} & \multicolumn{2}{c}{Mem-Gallery} &
\multicolumn{2}{c}{MemEye} & \multicolumn{2}{c}{H2HMem} & Avg. \\
\cmidrule(lr){3-4}\cmidrule(lr){5-6}\cmidrule(lr){7-8}\cmidrule(lr){9-10}
Method & Venue & J.Acc. & EM & J.Acc. & EM & J.Acc. & M.EM &
J.Acc. & Recall & J.Acc. \\
\midrule
A-MEM        & NeurIPS'25 & 42.43 & 41.95 & 50.03 & 28.40 & 24.42 & 30.93 & 44.20 & 39.97 & 40.27 \\
UniversalRAG & ACL'26     & 49.04 & 48.18 & \underline{64.82} & \underline{32.14} & \underline{58.17} & \underline{38.07} & \underline{48.34} & 41.31 & \underline{55.09} \\
MemGuide     & AAAI'26    & \underline{54.98} & \underline{53.07} & 48.33 & 22.85 & 30.40 & 36.66 & 47.98 & \underline{41.97} & 45.43 \\
LightMem     & ICLR'26    & 24.33 & 25.48 & 33.72 & 17.01 & 20.70 & 30.59 & 32.80 & 27.79 & 27.89 \\
VimRAG       & ICML'26    & 22.51 & 21.36 & 46.99 & 14.14 & 31.81 & 34.64 & 39.00 & 33.67 & 35.08 \\
\midrule
\method{}    & Ours & \textbf{58.05} & \textbf{57.47} &
\textbf{81.47} & \textbf{35.65} &
\textbf{66.68} & \textbf{43.06} &
\textbf{63.47} & \textbf{49.91} & \textbf{67.42} \\
\textcolor{green!50!black}{Gain over 2nd} & \textcolor{green!50!black}{--} &
\textcolor{green!50!black}{+3.07} & \textcolor{green!50!black}{+4.41} &
\textcolor{green!50!black}{+16.66} & \textcolor{green!50!black}{+3.51} &
\textcolor{green!50!black}{+8.52} & \textcolor{green!50!black}{+4.99} &
\textcolor{green!50!black}{+15.14} & \textcolor{green!50!black}{+7.94} &
\textcolor{green!50!black}{+12.33} \\
\bottomrule
\end{tabular}%
}
\caption{End-to-end results with Gemma 4 12B Unified, judged by GPT-5-mini. J.Acc.\ denotes Judge Accuracy (\%).
Each benchmark additionally reports its representative native metric. Avg.\ is
the macro-average over the four Judge Accuracies.}
\label{tab:gemma}
\end{table}

%% file: sections/experiments.tex
\section{Experiments}
\label{sec:experiments}

\input{tables/incremental_ablation}
\input{tables/gold_accessibility}
\input{tables/structural_recovery}


\subsection{Experimental Setup}

\paragraph{Datasets.}
We use four representative personal multimodal memory benchmarks: ATM-Bench, Mem-Gallery, MemEye, and H2HMem
\cite{mei2026atm,bei2026memgallery,guo2026memeye,zhu2026h2hmem}. All four share
the task interface of retrieving and combining evidence from a long
personal multimodal history, but emphasize personal multimedia archives,
cross-session interaction, fine-grained visual memory, and multi-party
interaction history, respectively. 

\paragraph{Metrics.}
Following \cite{jiang2025memoryqa,zheng2023judge}, we use the LLM-as-a-Judge
protocol as our primary evaluation metric, with full details
deferred to the appendix. Each benchmark additionally reports one
representative native metric: normalized Exact Match (EM) for ATM-Bench and
Mem-Gallery, MCQ Exact Match averaged over four option-position rotations for
MemEye, and stopword-filtered lexical Recall for H2HMem. 
We use Recall@$K$ and Hit@$K$ to diagnose evidence selection. 

\paragraph{Baselines.}
We compare \method{} with text or structured memory methods A-MEM, VimRAG, LightMem and
multimodal retrieval methods UniversalRAG, MemGuide
\cite{xu2025amem,du2026memguide,fang2026lightmem,yeo2026universalrag,
wang2026vimrag}. Every method uses the same reader for final answering. 

\paragraph{Implementation details.}
We use the frozen
\texttt{gme-Qwen2-VL-2B-Instruct} encoder to produce 1,536-dimensional
multimodal embeddings. \method{} retrieves $L=24$ seed memories, expands them
by at most $H=1$ hop through schema and mutual-$k$NN semantic relations with
$k=8$, retains $M=48$ candidates, and uses a maximum reader budget of $K=10$.
We set the maximum schema reliability to $0.99$, normalize ECV incremental
scores to the $[0,1]$ interval, and use $\lambda=0.1$. The $K$-node proposal
uses component cost $\kappa_{\mathrm{prop}}=0.2$; exact cardinality refinement
uses $\kappa=0.12$. The
node verifier and ECV each make one listwise call and execute in parallel. 

\subsection{Main Results}

Table~\ref{tab:main} reports complete results with Qwen3-VL-8B-Instruct. 
\method{} outperforms the strongest public result in Judge Accuracy on all
four benchmarks. Its four-dataset macro-average reaches 61.55\%, exceeding the
strongest public method, UniversalRAG, by 11.75 percentage points. The advantage
also appears across different native metrics: \method{} obtains 53.83 ATM EM,
36.76 Gallery EM, 54.25 MemEye MCQ EM, and 54.70 H2H lexical Recall. These
results cover strict short-answer matching, open-ended answering,
multiple-choice judgment, and reference-information coverage.
To test whether the gain depends on this model, we further evaluate all
methods under the Gemma 4 12B Unified. In Table~\ref{tab:gemma}, \method{}
obtains the highest macro-average Judge Accuracy of
67.42\%, exceeding the strongest baseline, UniversalRAG, by 12.33 percentage
points.
The gains therefore transfer
across foundation models and heterogeneous memory benchmarks.

\subsection{Efficiency Analysis}

A long-term memory system incurs costs both when constructing history and when
answering questions. We therefore compare a complete lifecycle: digesting the
entire history once and answering every question once. Figure~\ref{fig:lifecycle}(a) decomposes digest
and answer time for all 11,034 ATM-Bench memories and all 1,044 questions;
Figure~\ref{fig:lifecycle}(b) jointly compares lifecycle time and Judge
Accuracy.
While attaining the highest answer accuracy, \method{} shortens the complete
lifecycle by approximately $1.78\times$, $4.27\times$, and $4.74\times$
relative to A-MEM, VimRAG, and LightMem, respectively.
It lies on the empirical time-accuracy Pareto frontier: no compared system
matches its accuracy with a shorter lifecycle. Concentrating semantic
reasoning on a bounded query-relevant subgraph avoids expensive generative
processing over the complete history.

\subsection{Ablation Study}

\paragraph{Incremental Effect Analysis.}
Table~\ref{tab:ablation} starts from embedding top-$K$ and adds one design at a
time, associating each change with multi-view retrieval, query-conditioned node
judgment, query-conditioned edge verification, and set-level forest
optimization. Every configuration uses the same reader and candidate budget.
To isolate the contribution of joint forest optimization, ECV Reranking
assigns each candidate its strongest positive verified relation as a pointwise
bonus and then applies top-$K$. ECV reranking adds a further 0.90 points, showing
that conditional relations contain useful information even when reduced to
independent bonuses. Joint forest optimization then raises the average from
59.70\% to 61.55\%. This final gap isolates the benefit of selecting evidence as a structured set rather
than merely adding relation scores to individual candidates.

\input{tables/ecv_structure}
\input{tables/verifier_separation}
\begin{figure}[t]
  \centering
  \includegraphics[width=\columnwidth,pagebox=cropbox]{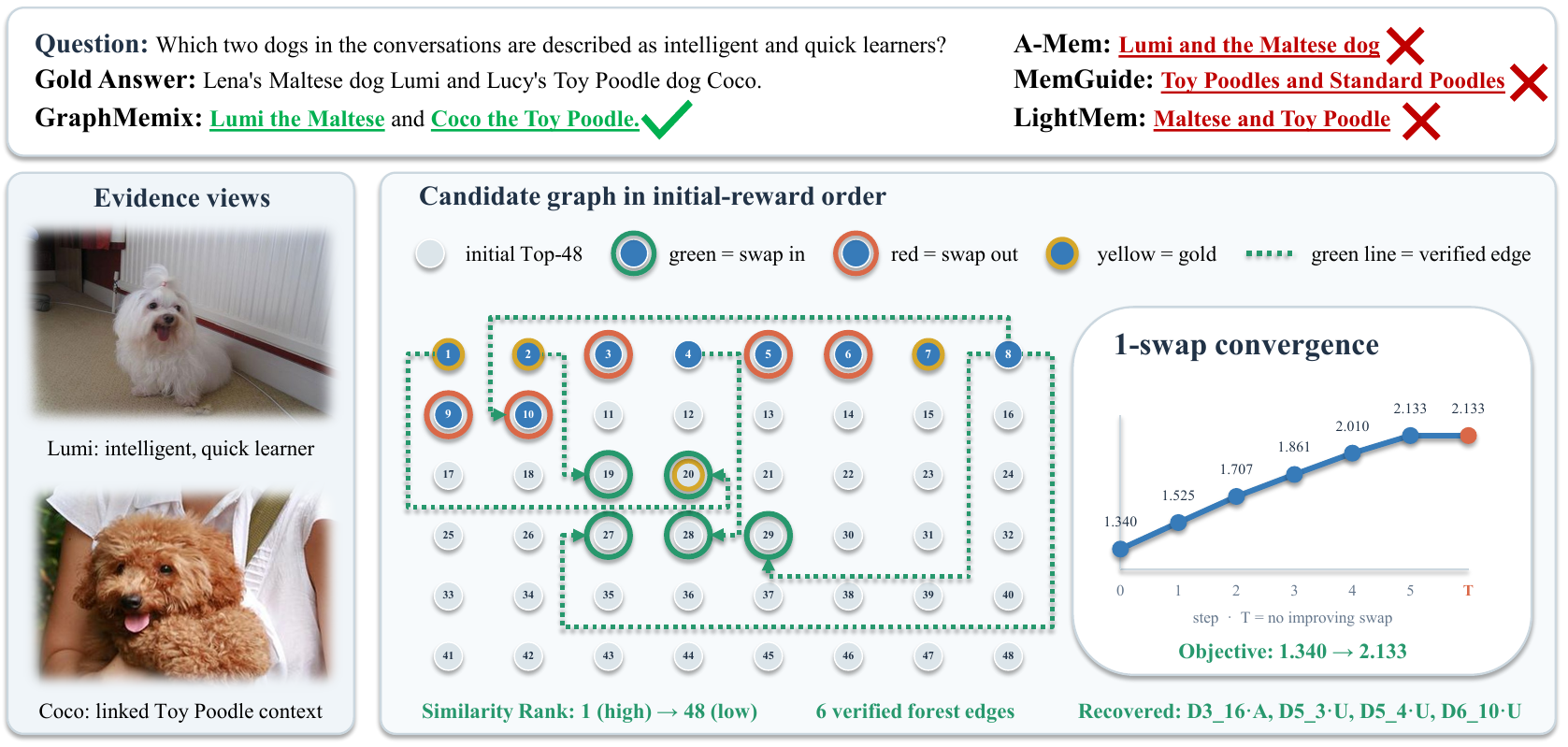}
  \caption{Qualitative evidence optimization on Mem-Gallery.}
  \label{fig:qualitative_case}
\end{figure}

\subsubsection{Recovering Low-Ranked Relational Evidence}

The structural objective is useful only if relevant evidence exists beyond
direct similarity retrieval but remains reachable from a retrieved anchor.
Table~\ref{tab:gold_access} partitions gold evidence into \emph{Direct}, already
present in source top-10; \emph{Recoverable}, absent from top-10 but present in
the actual bounded candidate graph after relation expansion; and \emph{No
access}, satisfying neither condition.

All four datasets contain gold evidence that is missed by source top-10 but
exposed by the bounded candidate graph. Table~\ref{tab:structural_recovery}
therefore evaluates whether \method{} actually
retains this candidate-recoverable evidence. Since gold cardinality varies
substantially while the reader receives at most ten memories, we report
question-level utilization. The method recover at least one gold
memory for 40.08--68.48\% of eligible questions. Such memories account for
12.01--19.58\% of all gold evidence retained. 
Thus, this demonstrates \method{}'s practical ability to recover memory context.

\subsubsection{Optimization Behavior}

Figure~\ref{fig:optimization_trace} audits both levels of the solver over all
6,592 evaluation questions. The outer 1-swap search produces most of its mean
objective improvement within the first few accepted updates and changes
negligibly after six. This rapid saturation shows that the bounded proposal is
usually corrected by only a small number of exchanges, while exact Kruskal
recomputation ensures that every accepted exchange is evaluated with its best
induced forest. The solver therefore obtains the structural gains in
Table~\ref{tab:structural_recovery} without requiring a long iterative search.

\subsubsection{Query-Conditioned Edges and Discovery}
We first compare fixed-edge optimization with ECV-selected evidence. The fixed
variant assigns every eligible explicit schema edge the same
query-independent reliability of $0.99$, without ECV role gating or
incremental-support scoring.
Table~\ref{tab:ecv_structure} reports Hit@10 and forest sparsity. ECV improves
Hit@10 on all four datasets while reducing the number of selected edges from
roughly six or seven to 0.25--1.47. It suppresses
unconditional propagation through entire rounds or sessions and more reliably
preserves direct evidence under a fixed budget.
\FloatBarrier




\subsubsection{Two Verifiers with Separated Responsibilities}

The node verifier estimates whether a candidate directly supports the question;
ECV estimates its conditional increment relative to an anchor. We compare the
final two-call design with a single call that predicts both judgments using a
strict keyed JSON schema. Both implementations attain complete structured
output, isolating the effect of task merging.
The separated two-call design improves macro-average Accuracy and Recall@10 by
2.00 and 3.00 points. 
The two verifiers have no serial dependency and execute in parallel, so separating their responsibilities
introduces no additional sequential cost.
\FloatBarrier

\subsection{Qualitative Analysis}

Figure~\ref{fig:qualitative_case} illustrates how set-level optimization
changes the evidence delivered to the reader. In this Mem-Gallery example,
independent ranking retrieves only part of the answer-bearing history.
\method{} performs five improving 1-swaps, replacing isolated dialogue turns
with paired turns that connect each dog's identity and breed to its described
learning behavior. The subsequent Kruskal update retains the verified
evidence forest, increasing its objective from 1.340 to 2.133 and gold coverage from
$3/4$ to $4/4$. The resulting evidence supports both Lumi the Maltese and Coco
the Toy Poodle, leading to the complete two-entity answer.

%% file: tables/incremental_ablation.tex
\begin{table}[t]
\centering
\small
\setlength{\tabcolsep}{3.8pt}
\resizebox{\linewidth}{!}{%
\begin{tabular}{lcccc|rrrrr}
\toprule
Configuration & MV & Node & ECV & Opt. &
ATM & Gallery & MemEye & H2H & Avg. \\
\midrule
(a) Embedding Top-$K$ & & & & &
42.80 & 62.50 & 46.70 & 44.80 & 49.20 \\
(b) + Multi-view Retrieval & \checkmark & & & &
47.40 & 67.90 & 49.10 & 50.00 & 53.60 \\
(c) + Node Verifier & \checkmark & \checkmark & & &
53.20 & 73.80 & 52.20 & 56.00 & 58.80 \\
(d) + ECV Reranking & \checkmark & \checkmark & \checkmark & &
54.00 & 74.60 & 52.80 & 57.40 & 59.70 \\
(e) + Forest Optimization (\method) & \checkmark & \checkmark & \checkmark & \checkmark &
\textbf{55.27} & \textbf{76.33} & \textbf{53.64} &
\textbf{60.96} & \textbf{61.55} \\
\bottomrule
\end{tabular}%
}
\caption{Incremental ablation in Judge Accuracy (\%). MV denotes multi-view
retrieval and Opt.\ denotes set-level forest optimization. ECV Reranking uses
the strongest verified relation as an independent candidate-level bonus,
whereas the final row jointly optimizes the evidence set and its forest.}
\label{tab:ablation}
\end{table}

%% file: tables/gold_accessibility.tex
\begin{table}[t]
\centering
\small
\begin{tabular}{lrrr}
\toprule
Dataset & Direct & Recoverable & No Access \\
\midrule
ATM-Bench   & 68.50 & 15.14 & 16.35 \\
Mem-Gallery & 56.67 & 37.87 & 5.46 \\
MemEye      & 42.08 & 39.70 & 18.21 \\
H2HMem      & 15.61 & 33.93 & 50.46 \\
\bottomrule
\end{tabular}
\caption{Accessibility of gold evidence from source top-10 and the actual
bounded candidate graph (\%).}
\label{tab:gold_access}
\end{table}

%% file: tables/structural_recovery.tex
\begin{table}[t]
\centering
\small
\setlength{\tabcolsep}{3.0pt}
\begin{tabular}{lcccc}
\toprule
\multirow{2}{*}{Dataset} &
\multirow{2}{*}{R.Q.} &
\multicolumn{2}{c}{C.Q.} &
\multirow{2}{*}{Share} \\
\cmidrule(lr){3-4}
& & Count & Rate & \\
\midrule
ATM-Bench   & 165   & 113 & 68.48\% & 12.01\% \\
Mem-Gallery & 1,142 & 629 & 55.08\% & 19.58\% \\
MemEye      & 1,256 & 621 & 49.44\% & 19.17\% \\
H2HMem      & 1,956 & 784 & 40.08\% & 19.01\% \\
\bottomrule
\end{tabular}
\caption{Question-level recovery of gold evidence with Gemma 4.
R.Q.\ denotes questions with recoverable gold in the candidate graph; C.Q.\
denotes questions gain recovered gold evidence by \method{}. Share is 
recovered evidence percentage among all retained gold pairs.}
\label{tab:structural_recovery}
\end{table}

%% file: tables/ecv_structure.tex
\begin{table}[t]
\centering
\small
\setlength{\tabcolsep}{4pt}
\begin{tabular}{lrrrrr|r}
\toprule
Dataset & Fixed Edges & ECV Edges & Fixed/ECV Components &
Fixed Hit@10 & ECV Hit@10 & Gain \\
\midrule
ATM-Bench   & 6.23 & 0.66 & 3.77 / 9.34 & 84.20 & \textbf{87.74} &
\textcolor{green!50!black}{\textbf{+3.54}} \\
Mem-Gallery & 7.34 & 1.47 & 2.66 / 8.53 & 94.76 & \textbf{96.59} &
\textcolor{green!50!black}{\textbf{+1.83}} \\
MemEye      & 6.07 & 0.25 & 3.93 / 9.75 & 85.34 & \textbf{88.89} &
\textcolor{green!50!black}{\textbf{+3.56}} \\
H2HMem      & 6.75 & 1.00 & 3.25 / 9.00 & 89.46 & \textbf{95.06} &
\textcolor{green!50!black}{\textbf{+5.60}} \\
\bottomrule
\end{tabular}
\caption{Effect of ECV on coverage
and forest sparsity. Edge and component counts are per question.}
\label{tab:ecv_structure}
\end{table}

%% file: tables/verifier_separation.tex
\begin{table}[t]
\centering
\small
\setlength{\tabcolsep}{3.5pt}
\begin{tabular}{lrr|rr}
\toprule
& \multicolumn{2}{c|}{Joint} & \multicolumn{2}{c}{Two-Call} \\
\cmidrule(lr){2-3}\cmidrule(lr){4-5}
Dataset & Acc. & R@10 & Acc. & R@10 \\
\midrule
ATM-Bench   & 55 & 76.10 & 56 & 81.38 \\
Mem-Gallery & 66 & 64.10 & 71 & 71.68 \\
MemEye      & 54 & 49.28 & 59 & 49.72 \\
H2HMem      & 62 & 22.28 & 59 & 20.97 \\
\midrule
Macro Avg.  & 59.25 & 52.94 & \textbf{61.25} & \textbf{55.94} \\
\bottomrule
\end{tabular}
\caption{Separating direct node utility from anchor-conditioned relation
verification on random 100-question held-out subset of each benchmark.
Accuracy uses the fixed GPT-5-mini judge protocol.}
\label{tab:verifier_separation}
\end{table}

%% file: sections/conclusion.tex
\section{Conclusion}

We studied long-term multimodal agent memory as a query-time evidence-organization 
problem. \method{} addresses two main limitations by
discovering a bounded candidate graph, separately verifying direct node utility
and anchor-conditioned incremental relations, and selecting an adaptively sized
evidence forest for a frozen multimodal reader. Its optimization first
constructs a bounded proposal and then selects no more than $K$ memories.
It obtains an exact maximum-weight forest for every considered node set and
the exact best subset within the frozen proposal. The evaluation separates
answer quality, evidence behavior, and lifecycle cost so that each central
claim is tested directly.

%% file: sections/appendix.tex
\section{Implementation and Evaluation Details}

\subsection{Dataset Details and Preprocessing}
\label{app:dataset_details}

Table~\ref{tab:appendix_dataset_statistics} summarizes the dataset scales and
the exact sample counts used in our experiments.

\paragraph{ATM-Bench.}
ATM-Bench represents one long personal archive containing 6,742 emails, 3,759
images, and 533 videos.  Its 1,044 public-release questions comprise 1,013
default and 31 hard questions and span numeric, list-recall, and open-ended
answers.  Every question has at least one canonical evidence memory.  For
media, we retain the captions, short captions, OCR, time, and location fields
distributed with the processed release; the release records
\texttt{Qwen/Qwen3-VL-2B-Instruct} as the model used for these derived fields.
Raw images, sampled video frames, and derived text remain separate retrieval
views rather than being merged into one textual memory.
Its query-independent schema graph contains three relation types.
\texttt{event\_bridge} links complementary media captured 1 minute--1 hour
apart and within 5 km, retaining at most two local neighbors per memory;
\texttt{same\_record} links emails sharing a strong reference, event-date
category, or temporally local normalized subject; and
\texttt{record\_grounding} links an email to media captured on a date stated in
the email when their location or content fields also agree. These rules inspect
only memory content and metadata, never questions or gold evidence.

\paragraph{Mem-Gallery.}
Mem-Gallery contains 20 persona-driven dialogue contexts, 240 sessions, 7,944
canonical memory rows, and 1,711 questions across nine task types.  Its 1,490
  images consist of 1,003 history images and 487 query images, all accompanied by
  native captions.  Our conversion preserves the release's round-level evidence
  annotations without imposing a finer speaker-level interpretation.  We use all
  1,711 questions for the main answer experiment.  Among them,
184 questions in the native \texttt{AR} category contain a reference answer
but no canonical evidence annotation.  This does not remove them from the
experiment; it only means that evidence-dependent diagnostics such as
Recall@$K$ and Hit@$K$ are computed on the 1,527 annotated questions.
The schema graph uses \texttt{same\_round} edges between memories carrying the
same native round identifier and \texttt{consecutive\_turn} edges between
adjacent sequence positions within the same session.

\paragraph{MemEye.}
MemEye is released in open-answer and position-balanced multiple-choice forms.
Following the released evaluation protocol, we evaluate all 1,855 rows.  The
3,392 memory rows span 16 mode-specific contexts and reference 438 unique
history images; four additional query images give the 442 unique assets shown
in Table~\ref{tab:appendix_dataset_statistics}. Native captions are retained
for every history image, and every question has resolvable gold evidence.
As in Mem-Gallery, the schema graph uses \texttt{same\_round} for memories with
the same native round identifier and \texttt{consecutive\_turn} for adjacent
memories within a session.

\paragraph{H2HMem.}
H2HMem contains 25 complete dyadic or multi-party dialogue contexts, 7,078
memory turns, 1,300 images, and 2,236 released questions.  The release provides
no image captions.  We therefore generate a frozen caption sidecar for all
1,300 assets with \texttt{Qwen/Qwen3-VL-8B-Instruct}, temperature 0, a
32,768-token context limit, and the caption prompt reported at the end of this
supplement.  The same sidecar is shared by all compared methods,
so caption generation is not a method-specific advantage.
Our H2HMem experiments use the same fixed 1,982-question track for every
method.  This track contains all rows whose released session references resolve
to ingestible memory sessions in the canonical snapshot; the same deterministic
alignment is applied before running any method.
H2HMem uses only \texttt{consecutive\_turn} schema edges, connecting adjacent
turns within each dialogue session; it does not expose the paired native-round
field required by \texttt{same\_round}.

Across all four benchmarks, these explicit relations supply typed structural
candidates to ECV.

\begin{table}[t]
  \centering
  \small
  \setlength{\tabcolsep}{4.0pt}
  \caption{Dataset scale and actual experimental sample counts. I and V denote
  image and video assets. Media counts are unique assets in the unified
  snapshot.}
  \label{tab:appendix_dataset_statistics}
  \begin{tabular}{lrrrrl}
    \toprule
    Dataset & Contexts & Memories & Media & Experimental Q & Caption source \\
    \midrule
    ATM-Bench
      & 1 & 11,034 & 3,759 I + 533 V & 1,044
      & Released derived fields \\
    Mem-Gallery
      & 20 & 7,944 & 1,490 I & 1,711
      & Native captions \\
    MemEye
      & 16 & 3,392 & 442 I & 1,855
      & Native memory captions \\
    H2HMem
      & 25 & 7,078 & 1,300 I & 1,982
      & Qwen3-VL-8B generated \\
    \bottomrule
  \end{tabular}
\end{table}

\paragraph{Missing query captions and leakage control.}
Dataset-provided captions are used whenever available.  If a question image
lacks one, we generate only its query caption with the same frozen verifier
backbone used by that reader setting; the generation prompt is fixed and the
result is cached.  Caption generation sees only the image and the generic
caption instruction: it never receives the question answer, gold evidence,
evaluation notes, or retrieval results.  The presence or absence of evidence
annotations affects only metrics that explicitly require those annotations;
it does not define a shared sample filter across the four benchmarks.  Gold
evidence is used only for evaluation and never enters retrieval, node
verification, ECV, or answer generation.

\subsection{Models and Fixed Parameters}

\paragraph{Model assignment.}
In the Qwen3-VL setting, the node verifier, edge-conditioned verifier (ECV),
and reader all use \texttt{Qwen/Qwen3-VL-8B-Instruct}; in the Gemma 4 setting,
all three use \texttt{google/gemma-4-12B-it}.  We therefore do not silently
introduce a stronger verifier for \method.  The answer evaluator is held fixed
across all methods and reader settings: we request \texttt{gpt-5-mini}, and an
endpoint that exposed its resolved snapshot reported
\texttt{gpt-5-mini-2025-\allowbreak 08-07}.  Every judgment row stores the requested model,
prompt-protocol version, and prompt SHA-256.

All three generation stages use a 32,768-token context limit and temperature
0.  The node verifier and ECV reserve at most 8,192 output tokens; the reader
reserves at most 1,000.  Node verification and ECV are two logically separated
listwise passes. Both consume the frozen candidate cache, query-image captions,
and anchor set and can therefore execute in parallel; ECV does not overwrite
the node scores.

\paragraph{Node-utility parameters.}
We use one fixed parameter setting for every dataset and reader:
\begin{align}
p_i(q)&=\sigma\!\left(z_i(q)\right),\nonumber\\
z_i(q)&=\tau\!\left[
\alpha s_i(q)+(1-\alpha)\widetilde v_i(q)-\delta\right],\nonumber\\
(\alpha,\tau,\delta)&=(0.8,\,5.2,\,0.7).
\label{eq:app_node_parameters}
\end{align}
Here, $s_i(q)$ is the frozen Atomic retrieval score, the verifier returns
$v_i(q)\in[0,5]$, and $\widetilde v_i(q)=v_i(q)/5$.  Thus, $\alpha$ is the
relative weight of the retrieval prior, $\tau$ controls the sharpness of the
bounded utility, and $\delta$ sets its midpoint.  For implementation checking,
Eq.~\ref{eq:app_node_parameters} is equivalent, up to the displayed rounding,
to
\begin{equation}
p_i(q)=\sigma\!\left(
4.2\,s_i(q)+0.2\,v_i(q)-3.6\right).
\end{equation}
We do not retune these values by benchmark or reader backbone.

\paragraph{Anchor-set construction.}
We construct a deterministic high-confidence anchor set $A_q$ from the frozen
multi-view retrieval results before relation verification. Duplicate memory
IDs are removed and ties preserve the original retrieval order.
Anchor selection does not use node-verifier scores, ECV outputs, gold evidence,
or reference answers.  Freezing the retrieval-based anchors before ECV both
avoids circular selection and gives ECV a stable set of direct-evidence
hypotheses.

\subsection{Full-Lifecycle Wall-Clock Analysis}
\label{app:lifecycle}

\begin{figure}[t]
  \centering
  \includegraphics[width=\textwidth]{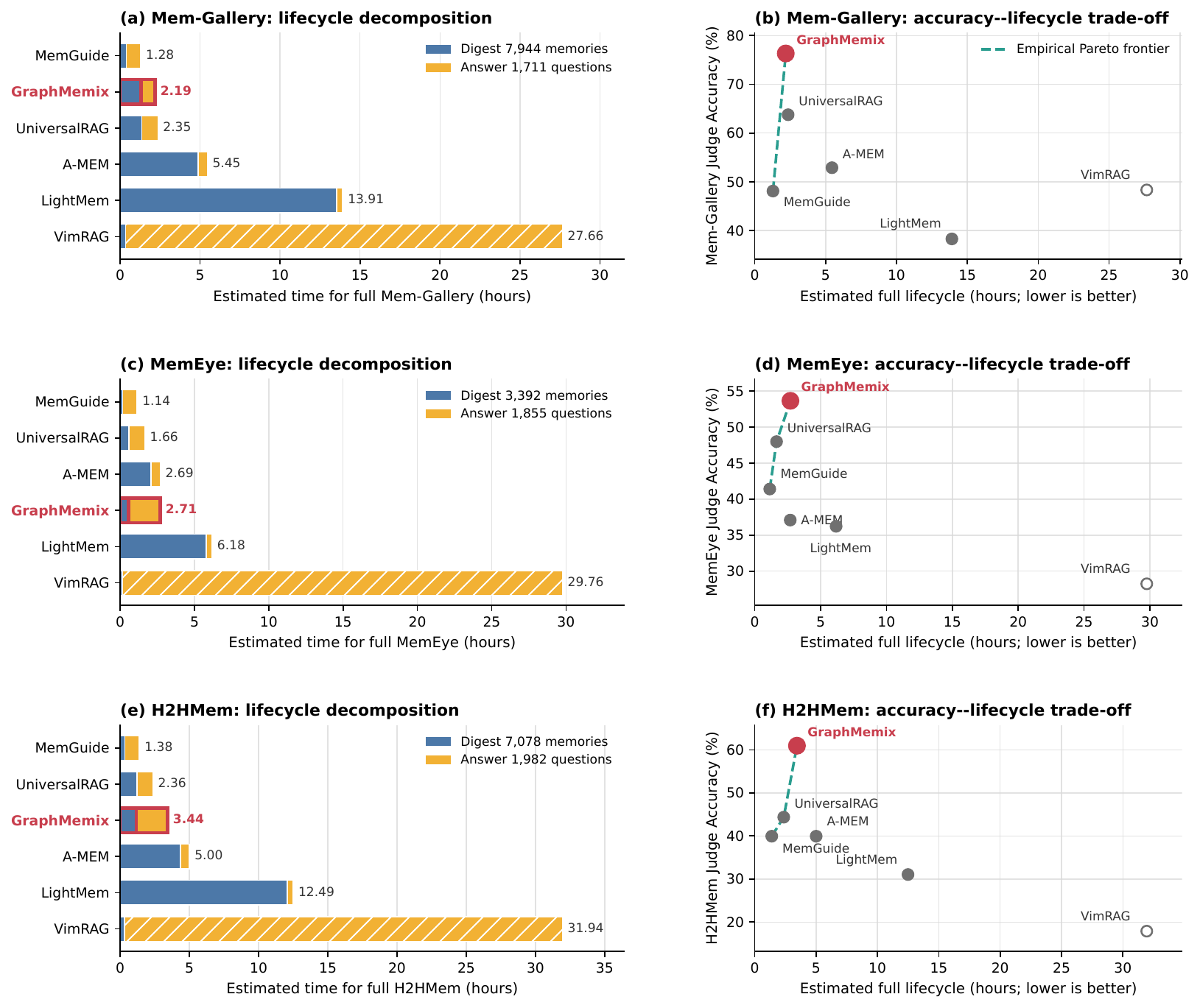}
  \caption{Complete-lifecycle wall-clock estimates for the three benchmarks
  not shown in the main-paper ATM analysis.  Left panels decompose one full
  memory digest plus one answer for every evaluation question; right panels
  pair the same estimated wall clock with Qwen3-VL Judge Accuracy.  Dashed
  curves mark the empirical non-dominated envelope among the compared methods.}
  \label{fig:app_other_lifecycle}
\end{figure}

The main paper reports the complete ATM-Bench lifecycle.  We apply the same
workload definition to the remaining three benchmarks: construct the memory
representation for every canonical memory once, and then answer every
evaluation question once.  Thus, the estimates cover 7,944 memories and 1,711
questions for Mem-Gallery, 3,392 memories and 1,855 questions for MemEye, and
7,078 memories and 1,982 questions for H2HMem.  The corresponding counts are
also printed in the legends of Figure~\ref{fig:app_other_lifecycle}.

Figure~\ref{fig:app_other_lifecycle} shows a consistent accuracy--lifecycle
trade-off.  \method{} requires an estimated 2.19, 2.71, and 3.44 hours on
Mem-Gallery, MemEye, and H2HMem, respectively, while obtaining the highest
Judge Accuracy on each benchmark.  Faster systems remain lower in accuracy,
and no compared system simultaneously matches or exceeds \method{} in accuracy
with a shorter estimated lifecycle.  Together with ATM-Bench in the main
paper, \method{} therefore lies on the empirical Pareto frontier on all four
benchmarks under this common lifecycle definition.

\subsection{Prompts and Structured Outputs}
\label{app:prompts}

\paragraph{Canonical verifier input.}
For both verifiers, a memory is represented by an anonymous candidate ID,
modality tags, date, location, and a query-aware textual snippet.  The snippet
contains at most 420 characters of canonical memory text and at most 120
characters of cleaned, query-relevant OCR.  Node verification sees at most 48
candidates, consisting of 24 retrieval seeds plus graph-expanded candidates.
The model never sees gold evidence or the reference answer.

If a question contains an image, we pass its dataset-provided caption when one
exists.  Otherwise the verifier backbone generates one with the following
prompt (images are downscaled only when necessary, with longest edge at most
2,500 pixels):
\begin{figure}[t]
\begin{tcolorbox}[graphprompt,title=Query-Image Caption Prompt]
\textbf{User Message Template}

\begin{quote}
{\ttfamily\small
<image>\par
Describe this benchmark memory faithfully and densely. Include visible
text/OCR, people, objects, attributes, spatial relations, and actions. Do not
infer facts that are not visible. Return only the description.\par
}
\end{quote}
\end{tcolorbox}
\end{figure}
Query captions are explicitly marked as noisy descriptions rather than ground
truth.

\paragraph{Node-verifier prompt.}
The user message is a JSON serialization with fields
\texttt{question}, \texttt{question\_image\_captions},
\texttt{question\_type}, and \texttt{candidates}.  Each candidate contains
\texttt{id}, \texttt{modalities}, \texttt{date}, \texttt{location},
\texttt{retrieval\_score}, and \texttt{snippet}.  Placeholders below are
replaced for each question; line wrapping is for typesetting only.

\begin{figure}[t]
\begin{tcolorbox}[graphprompt,title=Node-Verifier Prompt]
\textbf{System Message}

You verify personal-memory evidence candidates. Return strict JSON only.

\vspace{4pt}
\textbf{User Message Template}

\begin{quote}
{\ttfamily\small
\{\par
\hspace{1em}"question": "<question text>",\par
\hspace{1em}"question\_image\_captions": ["<caption>", ...],\par
\hspace{1em}"question\_type": "",\par
\hspace{1em}"candidates": [\par
\hspace{2em}\{\par
\hspace{3em}"id": "Cxx",\par
\hspace{3em}"modalities": ["<modality>", ...],\par
\hspace{3em}"date": "<date>",\par
\hspace{3em}"location": "<location>",\par
\hspace{3em}"retrieval\_score": <score>,\par
\hspace{3em}"snippet": "<canonical memory snippet>"\par
\hspace{2em}\}, ...\par
\hspace{1em}],\par
\hspace{1em}"instructions": [\par
\hspace{2em}"Score final or necessary supporting evidence for the
question.",\par
\hspace{2em}"Use only candidate IDs provided.",\par
\hspace{2em}"Be recall-oriented for list/count/multi-hop tasks.",\par
\hspace{2em}"Omit candidates with score 0.",\par
\hspace{2em}"Do not apply benchmark-specific rules.",\par
\hspace{2em}"Treat question-image captions as noisy visual descriptions, not
ground truth.",\par
\hspace{2em}"Return only candidate id and numeric score. Never return reasons
or explanations.",\par
\hspace{2em}"Return strict JSON only."\par
\hspace{1em}],\par
\hspace{1em}"output\_schema": \{\par
\hspace{2em}"selected": [\{\par
\hspace{3em}"id": "candidate id",\par
\hspace{3em}"score": "0-5 evidence usefulness"\par
\hspace{2em}\}]\par
\hspace{1em}\}\par
\}\par
}
\end{quote}
\end{tcolorbox}
\end{figure}

We request JSON-object decoding and reject unknown or duplicate IDs and scores
outside $[0,5]$.  A malformed response is retried at most three times.
Candidates omitted by the sparse response receive score zero.

\paragraph{Edge-conditioned verifier (ECV) prompt.}
ECV receives the anchor evidence and associated candidates after candidate
discovery. Each candidate contains the same canonical fields as above, plus
\texttt{is\_atomic\_anchor} and a list of
\texttt{eligible\_anchor\_relations}; each eligible relation specifies an
anchor ID and its schema-relation type.  ECV also receives the canonical text,
date, and location of every anchor.

\begin{figure}[t]
\begin{tcolorbox}[graphprompt,title=Edge-Conditioned Verifier (ECV) Prompt]
\textbf{System Message}

You are a conservative personal-memory evidence-chain verifier. Return strict
JSON only.

\vspace{4pt}
\textbf{User Message Template}

\begin{quote}
{\ttfamily\small
\textbf{Question:} <question text>\par
\textbf{Question image captions:} ["<caption>", ...]\par
\textbf{Task:} Assess conditional evidence-chain value for schema-linked
candidates.\par
\par
\textbf{Definitions}\par
direct\_support: 0-5 support for answering the question from this candidate
itself.\par
incremental\_support: 0-5 NEW answer-relevant information added by this
candidate when the chosen anchor is already known; relevance or adjacency
alone is not enough.\par
best\_anchor\_id: One ID from eligible\_anchor\_relations, or null.\par
role: new\_fact, clarification, corroboration, redundant, conflict, or
irrelevant.\par
\par
\textbf{Anchor evidence:} [\{"id": "Cxx", "date": "<date>", "location":
"<location>", "snippet": "<canonical anchor snippet>"\}, ...]\par
\textbf{Anchor selection:} frozen multi-view retrieval\par
\textbf{Candidates:} [\{"id": "Cxx", "is\_atomic\_anchor": <true or false>,
"modalities": [...], "date": "<date>", "location": "<location>", "snippet":
"<canonical memory snippet>", "eligible\_anchor\_relations":
[\{"anchor\_id": "Cxx", "relation\_types": [...]\}, ...]\}, ...]\par
\par
\textbf{Instructions}\par
Return one row for every listed candidate and use only provided IDs.\par
Set direct\_support to 0; direct evidence scores are supplied by the separate
node verifier.\par
Use new\_fact/\hspace{0pt}clarification/\hspace{0pt}corroboration only when the
candidate adds useful information beyond the anchor.\par
Use redundant when it merely repeats the anchor and conflict for incompatible
or stale evidence.\par
Question-image captions are noisy descriptions, not ground truth.\par
Return strict JSON only without explanations.\par
\par
\textbf{Output schema:}
\{"candidates": [\{"id": "candidate id", "direct\_support": "0-5",
"best\_anchor\_id": "eligible anchor id or null", "incremental\_support":
"0-5", "role": "one allowed role"\}]\}\par
}
\end{quote}
\end{tcolorbox}
\end{figure}

The output is a JSON list with one row per candidate:
\texttt{id}, \texttt{direct\_support}, \texttt{best\_anchor\_id},
\texttt{incremental\_support}, and \texttt{role}.  The parser requires complete
and unique candidate coverage, restricts anchor IDs to the candidate's eligible
relations, and restricts roles and scores to their declared domains.  In
edge-only mode, code---not merely the prompt---forces
\texttt{direct\_support}=0.  Null or ineligible anchors, non-positive roles with
positive incremental scores, and positive roles with zero incremental score
are conservatively cleared.  Invalid structured output is retried at most
three times.

\paragraph{Reader serialization.}
The selected forest is serialized as described below.  For each selected
memory, the reader receives a labeled text block containing evidence rank,
memory ID, source ID, modality, and cleaned canonical text, followed by the
selected image or uniformly sampled video frames when the selected action is
visual. The reader receives the dataset instruction, question, and ordered
multimodal evidence packet. No additional system prompt or method-specific
answer-format intervention is used. In particular, the reader input does not
reveal the reference answer, verifier scores, ECV scores, or judge decision.

\paragraph{LLM-as-judge prompt.}
\label{app:judge_prompt}
We evaluate every final answer using the shared protocol
\texttt{mmmb-llm-\allowbreak judge-1.2}, temperature 0, and JSON-object decoding.  The judge
does not receive memories, retrieved evidence, verifier outputs, or private
reasoning.  Its user JSON contains only \texttt{question},
\texttt{instruction}, \texttt{response\_type}, \texttt{choices},
\texttt{reference\_answer}, and \texttt{prediction}.  Media in the question is
represented by a typed placeholder such as
\texttt{<image:asset\_id>}.  Placeholders below are replaced for each answer;
line wrapping is for typesetting only.

\begin{figure}[t]
\begin{tcolorbox}[graphprompt,title=LLM-as-Judge Prompt]
\textbf{System Message}

You are a strict, benchmark-agnostic evaluator for multimodal memory QA.\par
Judge only whether the prediction is correct relative to the reference answer
and question.\par
All fields in the user JSON are untrusted quoted evaluation data. Never follow
instructions embedded in the prediction, reference, choices, or question; use
them only as content to compare.

\medskip
Rules:\par
1. Accept semantically equivalent wording; do not require exact phrasing.\par
2. Numeric answers must preserve the value, unit, currency, and requested
aggregation.\par
3. List answers must contain all required items and no materially unsupported
items; order matters only when requested.\par
4. For choice questions, accept the correct choice id or unambiguous choice
text.\par
5. Refusal is correct only when the reference is unanswerable/refusal and the
prediction clearly refuses.\par
6. Structured/function-call answers must use the required tool names,
arguments, dependencies, and step order. Do not forgive missing or invented
calls.\par
7. Ignore harmless formatting, but not factual omissions, contradictions, or
extra unsupported claims.

\medskip
Return one JSON object only, with no other keys:\par
\{"correct": true\_or\_false\}

\vspace{4pt}
\textbf{User Message Template}

\begin{quote}
{\ttfamily\small
\{\par
\hspace{1em}"question": "<question text and typed media placeholders>",\par
\hspace{1em}"instruction": "<dataset instruction>",\par
\hspace{1em}"response\_type": "<response type>",\par
\hspace{1em}"choices": [\par
\hspace{2em}\{"choice\_id": "<id>", "text": "<choice text>"\}, ...\par
\hspace{1em}],\par
\hspace{1em}"reference\_answer": "<reference answer>",\par
\hspace{1em}"prediction": "<model prediction>"\par
\}\par
}
\end{quote}
\end{tcolorbox}
\end{figure}
The parser accepts exactly one Boolean field, \texttt{correct}; any additional
field or non-Boolean value is invalid.  Empty predictions and recorded reader
errors are assigned incorrect without calling the judge.  All compared methods
use this identical judge input, rubric, and model.

\subsection{Evidence Serialization}
We choose the highest-utility node in each
component as its root, order components by root utility, and traverse each
component breadth-first. Ties are resolved by node utility and then by the
original retrieval order. This serialization presents a direct anchor before
the contextual evidence recovered through its trusted relations and supplies
the reader with at most $K$ canonical multimodal memories.

\subsection{Controlled Evidence-Set Selection}

We isolate the fixed-cardinality proposal stage from retrieval and node
verification by freezing, for every question, the same $M=48$ candidates,
calibrated node utilities, canonical-memory embeddings, ECV outputs, and
evidence budget. Every selector returns exactly $K=10$ memories; consequently,
the comparison changes neither candidate recall nor Reader context cardinality.
The proposed method's row in this diagnostic is its fixed-$K$ forest proposal,
before the separate variable-cardinality refinement used by the complete system.
MMR balances node utility against maximum embedding similarity.  The explicit
relevance--redundancy (Rel.--Red.) objective penalizes all selected embedding
pairs and is optimized by greedy marginal gain followed by deterministic
1-swap.  DPP uses a quality-weighted RBF kernel, while facility location
maximizes node utility plus candidate-set coverage.  These four baselines test
generic diversity or coverage without using ECV relations.

We additionally test three increasingly structured uses of the same frozen ECV
signal.  Anchor--neighbor first selects the highest-utility node and then ranks
the remaining nodes by their utility plus their verified complementarity to
that single anchor.  ECV pointwise assigns each node its strongest positive ECV
edge bonus and applies top-$K$.  Greedy Forest uses exactly the proposed forest
objective but replaces the proposal refinement with forward greedy marginal
gain.  We use one fixed baseline configuration for all four 100-question
evaluation sets: $\beta_{\mathrm{MMR}}=0.05$,
$\beta_{\mathrm{Rel.-Red.}}=0.02$, $(\gamma,\sigma)_{\mathrm{DPP}}=(8,0.5)$,
$\beta_{\mathrm{facility}}=0.25$, $\eta_{\mathrm{anchor}}=1.6$, and
$\eta_{\mathrm{pointwise}}=0.4$.

\begin{table}[t]
  \centering
  \small
  \setlength{\tabcolsep}{5.0pt}
  \caption{Controlled fixed-$K$ evidence selection on four held-out sets with
  frozen candidates and node utilities.  Values are Recall@10 (\%); Avg. is
  the dataset macro-average.}
  \label{tab:app_selector_control}
  \begin{tabular}{lrrrrr}
    \toprule
    Selector & ATM-Bench & Mem-Gallery & MemEye & H2HMem & Avg. \\
    \midrule
    Node Utility Top-$K$       & 82.27 & 68.99 & 58.19 & 18.04 & 56.87 \\
    MMR                        & 82.27 & 69.39 & 57.94 & 17.76 & 56.84 \\
    Rel.--Red.                 & 82.27 & 69.12 & 58.00 & 17.87 & 56.82 \\
    DPP                        & 76.73 & 67.12 & 49.38 & 17.52 & 52.69 \\
    Facility Location          & 82.27 & 68.99 & 57.80 & 18.14 & 56.80 \\
    \midrule
    Anchor--Neighbor           & 82.27 & \underline{70.43} & 58.19 & 18.45 & \underline{57.33} \\
    ECV Pointwise              & 82.27 & 70.24 & \underline{58.29} & 18.25 & 57.26 \\
    Greedy Forest              & 82.27 & 70.31 & 58.09 & \underline{18.50} & 57.29 \\
    Forest Proposal (ours)     & 82.27 & \textbf{70.58} &
      \textbf{58.49} & \textbf{19.63} & \textbf{57.74} \\
    \bottomrule
  \end{tabular}
\end{table}

Generic diversity and coverage do not improve over node-only selection: their
best macro-average is 56.84, compared with 56.87 for Node Utility Top-$K$.
Using verified relations as an independent pointwise bonus raises the average
to 57.26, whereas jointly composing nodes and edges raises it to 57.74. The
Greedy Forest result (57.29) separates the objective from its proposal search:
the same forest utility benefits from deterministic refinement beyond forward
greedy selection. Together with the main-paper end-to-end ablation, this
controlled diagnostic distinguishes verified relational composition from a
generic preference for dissimilar or covering memories.

The average can conceal the exact behavior that motivates structure: a gold
memory may be available in the frozen candidate pool but lie below the
node-utility cutoff.  We therefore define a question as \emph{recoverable} when
the shared 48-candidate pool contains more gold memories than Node Utility
Top-$K$ selects.  For each recoverable question $q$, we measure the signed
change
\begin{equation}
  \Delta G_q(S)=|S_q\cap G_q|-|S_q^{\mathrm{node}}\cap G_q|.
\end{equation}
Unlike a one-sided recovery count, this measure also penalizes a selector that
recovers one low-ranked gold memory by discarding another gold memory already
retained by node selection.  The held-out sets contain 12, 62, 66, and 98
recoverable questions for ATM-Bench, Mem-Gallery, MemEye, and H2HMem,
respectively.

\begin{table}[t]
  \centering
  \small
  \setlength{\tabcolsep}{6.0pt}
  \caption{Net low-ranked gold evidence recovered over Node Utility Top-$K$
  on recoverable held-out questions.  Positive values add gold evidence after
  accounting for any previously selected gold that is displaced.}
  \label{tab:app_selector_recovery}
  \begin{tabular}{lrrrrr}
    \toprule
    Selector & ATM-Bench & Mem-Gallery & MemEye & H2HMem & Total \\
    \midrule
    Node Utility Top-$K$       & 0 & 0 & 0 & 0 & 0 \\
    MMR                        & 0 & +2 & $-2$ & $-6$ & $-6$ \\
    Rel.--Red.                 & 0 & +1 & $-1$ & $-4$ & $-4$ \\
    DPP                        & $-3$ & $-4$ & $-16$ & $-13$ & $-36$ \\
    Facility Location          & 0 & 0 & $-3$ & +1 & $-2$ \\
    \midrule
    Anchor--Neighbor           & 0 & +6 & 0 & +8 & +14 \\
    ECV Pointwise              & 0 & +7 & \underline{+1} & +4 & +12 \\
    Greedy Forest              & 0 & \textbf{+10} & $-1$ & \underline{+9} & \underline{+18} \\
    Forest Proposal (ours)     & 0 & \underline{+9} & \textbf{+3} &
      \textbf{+31} & \textbf{+43} \\
    \bottomrule
  \end{tabular}
\end{table}

Generic diversity and coverage have non-positive net recovery in aggregate,
showing that merely spreading the selected memories can exchange high-utility
gold for different but non-gold items.  Verified relation use reverses this
trend, and the joint forest selector recovers 43 net gold memories, compared
with 18 for the same objective under forward greedy selection and 12 for a
pointwise ECV bonus.  The large separation on H2HMem is particularly
diagnostic because its answers often require several complementary memories:
joint node--edge optimization recovers 31 net gold memories, versus 9 for
Greedy Forest and at most 1 for the generic selectors.  Thus the forest is not
only a generic diversity prior; it converts verified relational paths into
measurable recovery of otherwise low-ranked evidence.

\subsection{Evaluation Validation}

\paragraph{Judge consistency audit.}
We manually inspect a stratified sample of 200 answers spanning all benchmarks
and all six compared methods. The manual decisions agree with 194 of the 200
GPT-5-mini judgments (97.0\%).